\documentclass[12pt]{colt2019} 
 \pdfoutput=1
\usepackage{nicefrac}
\usepackage{caption}
\usepackage{booktabs}       
\title[Generalization Bounds for Autoencoders]{Generalization Bounds For Unsupervised and Semi-Supervised Learning With Autoencoders}
\usepackage{times}

\coltauthor{\Name{Baruch Epstein} \Email{baruch.epstein@gmail.com}\and
\Name{Ron Meir} \Email{rmeir@ee.technion.ac.il}\\
\addr Viterbi Faculty of Electrical Engineering, Technion - Israel Institute of Technology}

 \editor{Under Review for COLT 2019}

\begin{document}

\maketitle

\global\long\def\norm#1{\left\Vert #1\right\Vert }

\begin{abstract}%
  Autoencoders are widely used for unsupervised learning and as a regularization scheme in semi-supervised learning. However, theoretical understanding of their generalization properties and of the manner in which they can assist supervised learning has been lacking. We utilize recent advances in the theory of deep learning generalization, together with a novel reconstruction loss, to provide generalization bounds for autoencoders. To the best of our knowledge, this is the first such bound. We further show that, under appropriate assumptions, an autoencoder with good generalization properties can improve any semi-supervised learning scheme. We support our theoretical results with empirical demonstrations.
\end{abstract}


\section{Introduction}

An autoencoder (AE) (\cite{Hinton_Salakhutdinov_2006}) is a type of feedforward neural network, aiming at
reconstructing its own input through a narrow bottleneck. It typically
comprises two parts: $\text{enc}:\mathcal{X}\rightarrow\mathcal{R}$
and $\text{dec}:\mathcal{R}\rightarrow\mathcal{X}$ with $r\in\mathcal{R}$
some representation of the input $x\in\mathcal{X}$. Usually, $r$
is of a smaller dimension than $x$. The network is trained to find
encoder and decoder functions such that some loss $l\left(x,\text{dec}\left(\text{enc}\left(x\right)\right)\right)$
is minimized. A typical choice is the square loss $\norm{x-\text{dec}\left(\text{enc}\left(x\right)\right)}_{2}^{2}$.
In recent years, autoencoders have emerged as a standard tool
for both unsupervised and semi-supervised learning (SSL) (\cite{Denoising_AEs_10}, \cite{Zhuang_15}, \cite{ghifary2016deep}, \cite{DSN}, \cite{JAE_18}). Unfortunately, as
is frequently the case with deep learning approaches, the empirical
practice has not been matched by parallel advances in theory. That
is, unsupervised learning with autoencoders has not been able to benefit
from the recent bounds for supervised deep learning (\cite{Bartlett_Spectral_17}, \cite{Neyshabur_18}
\cite{Arora_Compression_18}, \cite{Shamir_Size_Independent_Bounds_18}). In SSL, the discrepancy is more severe still. There
is a fundamental tension between the goals of a supervised learner
and those of an AE. One might say that an AE ``wants to remember
everything a classifier wants to forget''. Indeed, given two slightly
different images of the digit $3$, a classifier would like to
ignore all differences, aiming instead to see them as similar objects.
An AE, on the other hand, aims at reconstructing precisely those nuances
(e.g. width, location in the image, style) that do not matter for
the classification. Some works (\cite{Bengio_Contractive_11}) have attempted
to encourage AEs to weigh classification-relevant features more heavily,
but this still dodges the basic question - why might an AE be useful
for supervised learning?

We address the gaps above. First, we introduce a margin-based reconstruction
loss which allows a natural adaptation of existing generalization bounds
for autoencoders, and show that a bounded loss in that sense implies a
bounded loss in the standard $L_{2}$ metric. Second, we demonstrate
a mechanism by which a well-performing autoencoder is likely to assist
in SSL; namely, it allows one to reduce dimension while \emph{preserving
the input structure}. More formally, we show that if the input distribution
satisfies a certain clustering assumption, then the encoder part of
an autoencoder with a small generalization error maps most of the
input to a low-dimensional distribution that itself satisfies a reasonably-good
clustering assumption. Finally, we extend \cite{Singh_08} by showing
conditions under which any supervised learner can benefit from AE-enabled
SSL.

The remainder of the paper is organized as follows. In Section \ref{sec:Related} we review some prior work on AEs and SSL. In \ref{sec_background}, we survey some recent generalization bounds for supervised
learning with deep networks, and the characterization by \cite{Singh_08} of conditions under which SSL is guaranteed to be beneficial.
In Section \ref{sec_generalization_bounds_AEs}, we introduce our proposed reconstruction loss and use
it to obtain generalization bounds for AEs. In Section \ref{sec_AEs_SSL}, we apply
these bounds to show that if an AE generalizes well, its encoder is limited in its ability to shrink the distances between most input pairs and discuss what
this implies for semi-supervised learning and \cite{Singh_08}. In Section \ref{sec_experiments}, we explore our bounds
empirically. Finally, we discuss possible implications of this work
and some future research directions.

The main contributions of the present work are the following. \emph{(i)} We adapt recent margin-based generalization bounds for feedforward networks to autoencoders via a novel loss. \emph{\emph(ii)} We tie good AE reconstruction performance to a non-contractiveness property of the encoder component. \emph{(iii)} We show that this implies the ability to trade off separation margins between input clusters for reduced dimension, which is beneficial for semi-supervised learning.  

\section{Related Work}\label{sec:Related}


A great deal of work has been devoted to dimensionality reduction since the introduction of (linear) principal component analysis (PCA) in 1901 by Karl Pearson. Nonlinear manifold-based methods introduced in \cite{roweis2000nonlinear,tenenbaum2000global}, were followed by work on AEs \cite{Hinton_Salakhutdinov_2006} that led to significantly improved results for practical problems. Later work by \cite{Denoising_AEs_10}, introduced a de-noising based criterion for training AEs, and demonstrated its improved representation quality compared to a reconstruction based criterion, contributing to better classification performance on subsequent supervised learning tasks. Further details, and a survey of AEs, can be found in \cite{bengio2013representation}.

Subsequent work directly addressed the SSL setting. Several papers demonstrated the empirical utility of SSL \cite{Bengio_Contractive_11,ranzato2008semi,rasmus2015semi,weston2012deep}. Within the related transfer learning setting, \cite{Zhuang_15} showed how to improve learning performance by combining two types of encoders, the first, based on an unsupervised embedding from the source and target domains, and the second, based on the labels available from the source data. \cite{ghifary2016deep}, suggested a joint encoder for both classification of labeled data, and reconstruction of unlabeled data, thereby maintaining both types of information, and enhancing performance in the face of scarce labels.  \cite{DSN} and \cite{JAE_18} use AEs for SSL and semi-supervised transfer learning, by explicitly learning to separate representations into private and shared components.

Within the framework of statistical learning theory, several recent papers have significantly improved previous generalization bounds for deep networks by incorporating more refined attributes of the network structure, aiming to explain the paradoxical effect of improved performance while over-training the network. Using covering number techniques, \cite{Bartlett_Spectral_17} provide margin based bounds that relate generalization error to the network's Lipschitz constant and matrix norms of the weights.   \cite{Neyshabur_18} establish similar matrix-norm-based margin bounds using a PAC-Bayes approach. \cite{Arora_Compression_18} present compression-based results by compressing the weights of a well performing network and bounding the error of the compressed network. Finally, \cite{Shamir_Size_Independent_Bounds_18} are able, under certain (restrictive) assumptions on matrix norms, to achieve generalization bounds that are completely independent of the network size. 

The value of SSL has been subject to much debate. \cite{Rigollet_07} provide a mathematical framework for the intuitive \emph{cluster assumption} of  \cite{Seeger_00}, and show that for unlabeled data to be beneficial, some clustering criterion is required (specifically, that the data consists of separated clusters with identical labels within each cluster). Based on a density level set approach, they prove fast rates of convergence in the SSL setting. \cite{Lafferty_07} and \cite{Niyogi_08}) study SSL within a minimax framework, the former work shows that, under the so-called manifold assumption, optimal minimax rates of convergence may be achieved, while the second work demonstrates a separation between two classes of problems. When the structure of the data manifold is known, fast rates can be achieved, while without such knowledge, convergence cannot be guaranteed. More directly related to our work, and building on the clustering assumption, \cite{Singh_08} identify situations in which semi-supervised learning can improve upon supervised learning. Unfortunately, their SSL bound suffers from the curse of dimensionality, and so depends exponentially on the dimension. Our work can be seen as allowing a trade-off between the clustering separation and the dimension, suggesting how to improve the bounds in \cite{Singh_08}. 

\cite{Rooyen_15} provide a principled approach to feature representation, and characterize the relation between the information retained by features about the input, and the loss incurred by a classifier based on these features. They suggest the application of their results to SSL, but do not provide explicit conditions or generalization bounds in this setting. Recently, \cite{supervised_AEs_18} provide such bounds for semi-supervised learning with linear AEs and a joint reconstruction/classification loss, using uniform stability arguments. They also provide empirical results that show that nonlinear AEs can indeed contribute to supervised learning. While their results are close in spirit to ours, we are more concerned with understanding how the structure of the data affects SSL, and, in particular, characterizing when, and to what extent, clustering of the input contributes to performance through unsupervised learning. Moreover, we rely on bounds that are specific to neural networks, rather than on the looser stability based bounds. 

\section{Background} \label{sec_background}
\subsection{Generalization for Feed-forward Networks} \label{generalization_for_NNs}

Let $\mathcal{X}$ be an input space, $\mathcal{Y}$ an output space, and $\mathcal{D}_{\mathcal{X},\mathcal{Y}}$ a distribution
on $\mathcal{X} \times \mathcal{Y}$. Throughout the paper, we shall assume that $\mathcal{X}\subset \mathbb{R}^N$ and that $\norm{x} \le B,\, \forall x\in \mathcal{X}$. Denote by $\mathcal{D}$ the marginal distribution on the inputs. The $j$-th entry of a vector $v$ is denoted by $v\left[j\right]$. For a collection of matrices $\boldsymbol{w}=\left\{ W_{i}\right\} _{i=1}^{d}$,
denote by $f_{\boldsymbol{w}}\left(x\right)$ the $d$-layer feedforward
network $W_{d}\phi\left(W_{d-1}\left(\phi...\phi\left(W_{1}x\right)\right)\right)$, where $\phi$ is a non-linearity. We shall focus on 	the ReLU function $\phi(x)=\max(0,x)$.
Let us recall a few recent results (\cite{Bartlett_Spectral_17}, \cite{Neyshabur_18},
\cite{Arora_Compression_18}) concerning supervised $K$-way classification with feedforward
networks. The output of a network $f_{\boldsymbol{w}}$ is a vector
$v\in\mathbb{R}^{K}$. For a pair $(x,y)$, define the (supervised) $\gamma$-margin loss as

\begin{align}
l_{\gamma}^{s}\left(f_{\boldsymbol{w}}(x),y\right) & \triangleq \begin{cases}
0 & f_{\boldsymbol{w}}\left(x\right)[y]-\gamma\ge\max_{j\neq y}f_{\boldsymbol{w}}\left(x\right)\left[j\right]\\
1 & \mathrm{o.w}.
\end{cases}.
\end{align}
That is, the loss is only $0$ if the correct, $y$-th output entry $f_{\boldsymbol{w}}\left(x\right)[y]$ is not only the largest one - but the second-largest entry is at least $\gamma$ away. Given a sample of size $m$, Denote by $\hat{L}^{s}_\gamma$ and $L^{s}_\gamma$ the corresponding empirical and expected function losses
\begin{align}
\hat{L}_{\gamma}^{s}\left(f_{\boldsymbol{w}}\right) & \triangleq \frac{1}{m}\sum_{i=1}^{m}l_{\gamma}^{s}\left(f_{\boldsymbol{w}}(x_{i}),y_{i}\right), \nonumber\\
L_{\gamma}^{s}\left(f_{\boldsymbol{w}}\right) & \triangleq {\displaystyle \mathbb{E}_{\left(x,y\right)\sim\mathcal{D}_{\mathcal{X},\mathcal{Y}}}l_{\gamma}^{s}\left(f_{\boldsymbol{w}}(x),y\right)}.
\end{align}
Note that $L^{s}_0$ is the standard $0-1$ classification loss. \\

All three aforementioned papers can be considered to suggest the same type
of claim,
\begin{align}
L_{0}^{s}\left(f\right)& \le\hat{L}^{s}_{\gamma}\left(f\right)+\Delta\left(f,m,\delta,\gamma\right), \mathrm{w.p}. \ge 1-\delta,
\end{align}
where $\Delta\left(f,m,\delta,\gamma\right)$ is a generalization
term depending on the network
parameters, failure probability $\delta$, sample size $m$ and margin
$\gamma$. We shall use the bound appearing in \cite{Neyshabur_18}, as it is the simplest to state\footnote{The bound in \cite{Bartlett_Spectral_17} is strictly tighter and allows for non-linearities other than ReLU,  however.}, but the similar results appearing in the other two papers can be adapted for our purpose as well.
\begin{proposition} \label{prop_original_Neyshabur}
(\cite{Neyshabur_18}) For any $0 < \delta < 1,0<\gamma$,
with probability at least $1-\delta$ over a training set of size
$m$, for any $f_{\boldsymbol{w}}$ of depth $d$ and a constant $C\left(B, f_{\boldsymbol{w}} \right)$ \footnote{See Sec. \ref{subsec_Neyshabur} for a more detailed bound statement.} depending only on the maximal input norm $B$ and on the structure of $ f_{\boldsymbol{w}}$, we have
\begin{align}
L_{0}^{s}\left(f_{\boldsymbol{w}}\right) & \le\hat{L}^{s} 	_{\gamma}\left(f_{\boldsymbol{w}}\right)+\text{\ensuremath{\mathcal{O}}}\left(\sqrt{\frac{C\left(B, f_{\boldsymbol{w}} \right)+\ln\frac{dm}{\delta}}{\gamma ^{2}m}}\right).
\end{align}
\end{proposition}
~

\subsection{Semi-Supervised Learning - Now It Helps Now It Doesn't}

We briefly review the necessary background from \cite{Singh_08},
stating the results and terminology in a somewhat simplified manner.
First, we define the clustering assumption they are working under.
Suppose the input distribution $\mathcal{D}$ is a finite mixture
of smooth component densities $\left\{ \mathcal{D}_{k}\right\} _{k=1}^{\mathcal{K}}$ with disjoint supports.
Suppose further that each $\mathcal{D}_{k}$ is bounded away from
zero and supported on a unique compact and connected set $C_{k}\subset\mathcal{X}$ with smooth boundaries:
\begin{align}
C_{k} & =\left\{ x\equiv\left(x_{1},...,x_{d}\right):g_{k}^{\left(1\right)}\left(x_{1},...,x_{d-1}\right)\le x_{d}\le g_{k}^{\left(2\right)}\left(x_{1},...,x_{d-1}\right)\right\} ,
\end{align}
where $g_{k}^{\left(1\right)},g_{k}^{\left(2\right)}$ are $(d-1)$-dimensional
Lipschitz functions. Finally, assume the target label is constant
on each $C_{k}$\footnote{Inputs with equal labels need not form a single cluster. Indeed,
typically they form a number of separate clusters.}. Then we say $\mathcal{D}$ satisfies the \emph{clustering assumption
with cluster-margin $\eta$}\footnote {The notation in the original is $\gamma$. We have changed it to avoid confusion with the $\gamma$-margin loss.} if each two clusters $C_{j},C_{k}$ are at
least $\eta$ apart. More formally, for $j,k\in\left\{ 1,..,K\right\} $,
let
\begin{align}
d_{jk} & =\begin{cases}
\displaystyle \min_{p,q\in\left\{ 1,2\right\} }\norm{g_{j}^{\left(p\right)}-g_{k}^{\left(q\right)}}_{\infty} & j\neq k\\
\, & \, \\
\norm{g_{k}^{\left(1\right)}-g_{k}^{\left(2\right)}}_{\infty} & j=k
\end{cases}.
\end{align}
Then the cluster-margin $\eta$ is simply ${\displaystyle \min_{j,k}}\,d_{jk}$. The support sets of the components $C_k$ in $\mathcal{D}$ are called \emph{the decision sets of $\mathcal{D}$}. Denote by $\mathcal{H}$ the set of all hypotheses $h:\mathcal{X}\rightarrow\mathcal{Y}$. 

\begin{definition}
A \emph{clairvoyant supervised learner} $\mathcal{A}_{\mathcal{D},n}:(\mathcal{X}\times \mathcal{Y})^n \rightarrow \mathcal{H}$  is a function mapping labeled training sets of size $n$ to hypotheses in $\mathcal{H}$, with perfect knowledge of the decision sets of $\mathcal{D}$. A \emph{semi-supervised learner} $\mathcal{A}_{m,n}:\mathcal{X}^m\times(\mathcal{X}\times \mathcal{Y})^n \rightarrow \mathcal{H}$  is a function mapping unlabeled training sets of size $m$ and labeled training sets of size $n$ to hypotheses in $\mathcal{H}$.
\end{definition}
The following theorem (a slightly weaker version of Corollary 1 in \cite{Singh_08})  asserts that under suitable conditions, semi-supervised
learning can always perform as well as any clairvoyant learner. 
\begin{proposition}(\cite{Singh_08}) \label{prop_Singh}
Let $\mathcal{D}$ satisfy the clustering assumption with cluster-margin $\eta$. Assume L is a bounded loss. Denote by $\mathcal{E}(\mathcal{A})=L(\mathcal{A}) - L^*$ the excess loss of a learner $\mathcal{A}$, where $L^*$ is the infimum loss over all possible learners. Suppose there exists a clairvoyant learner  $\mathcal{A}_{\mathcal{D},n}$ for which
\begin{align}
 \mathbb{E}[\mathcal{E}(\mathcal{A}_{\mathcal{D},n})] \le \varepsilon(n).
\end{align}
Then there exists a semi-supervised learner $\mathcal{A}_{m,n}$ such that if $\eta > C_0 ((\log m)^2 / m)^{1/N}$, then
\begin{align}
\mathbb{E}[\mathcal{E}(\mathcal{A}_{m,n})] \le \varepsilon(n) + O\left( \frac{1}{m}+n \left( \frac{(\log m)^2}{m} \right) ^{1/N} \right).
\end{align}
The constant $C_0$ does not depend on $\eta$, $m$ or $n$.

\end{proposition}

\begin{remark}
Note the exponential dependence on the input dimension $N$ in Prop. \ref{prop_Singh}. Mapping the input to a significantly lower dimension without decreasing $\eta$ too much is beneficial for the bound.
\end{remark}

\section{Generalization Bounds for Autoencoders} \label{sec_generalization_bounds_AEs}

Let us now turn to autoencoders and their generalization properties. We introduce a novel entry-wise $\gamma$-margin reconstruction loss and state a generalization bound for this loss. Furthermore, we show that such a bound implies a bound for the standard $L_2$ loss as well. \\

For simplicity, we consider $\mathcal{X}\in\left\{ 0,1\right\} ^{M}$.\footnote{All the definitions and results from here on can be extended straightforwardly to support finer input resolution, that is, to allow input values on a discrete grid $\left\{ 0,1/s,2/s,...,(s-1)/s,1\right\} $ for some integer $s$. The forms of the bound in Prop. \ref{thm_Neyshabur_AEs} and Thm. \ref{thm_mu_bound} do not change with $s$, though the $\gamma$-margin loss of any given autoencoder might. The $1/2$ value in the definition of $R(r, \gamma)$ in Sec. \ref{sec_AEs_SSL} is replaced by $s/2$. 
} We consider feed-forward fully-connected networks with output entries in $\left[0,1\right]$. \footnote{The restriction of the output to $[0,1]$ can be achieved by applying a sigmoid to the output, with the beneficial side effect of dividing the network Lipschitz constant by 4, as the Lipschitz constant of the sigmoid function is 1/4.}  Given a sample $x$ and a network $f_{\boldsymbol{w}}$,
the reconstructed output is $\hat{x}=f_{\boldsymbol{w}}\left(x\right)$,
though we will sometimes abuse the notation and simply write $f\left(x\right)$ or $f$.
Note that while the inputs are binary, the prediction for each entry
can be an intermediate value. An autoencoder network $f$ is a composition
of an encoder $enc$ and a decoder $dec$, both fully-connected feedforward networks.

\begin{definition}
For a margin $\gamma<\frac{1}{2}$, we define the $\gamma$-margin loss to
be the average amount of entries that were not reconstructed with a confidence of
at least $\gamma$. That is, 
\begin{align}
l_{\gamma}\left(x,\hat{x}\right) & :=
\frac{1}{M}{\displaystyle \sum_{j=1}^M\boldsymbol{1}\left(\left|x\left[j\right]-\hat{x}\left[j\right]\right|>\frac{1}{2}-\gamma\right)},
\end{align}
where $\boldsymbol{1}$ is the indicator function.
\end{definition}

Note that the loss is bounded between $0$ and $1$. The corresponding
expected loss and empirical loss on $m$ samples are denoted $L_{\gamma}$
and $\hat{L}_{\gamma}$, respectively:

\begin{align} \label{eq_gamma_rec_loss}
\hat{L}_{\gamma}\left(f_{\boldsymbol{w}}\right) & :=\frac{1}{m}\sum_{i=1}^{m}l_{\gamma}\left(x_{i},f_{\boldsymbol{w}}(x_{i})\right) \nonumber\\
L_{\gamma}\left(f_{\boldsymbol{w}}\right) & :={\displaystyle \mathbb{E}_{x\sim\mathcal{D}}l_{\gamma}\left(x,f_{\boldsymbol{w}}(x)\right)}.
\end{align}
We can adapt Prop. \ref{prop_original_Neyshabur} in Sec. \ref{generalization_for_NNs} for autoencoders and the losses defined in Eq. \ref{eq_gamma_rec_loss}. 
\begin{theorem} \label{thm_Neyshabur_AEs} For any positive $0<\delta<1,0<\gamma_1 < \gamma_2<1/2$,
with probability at least $1-\delta$ over a training set of size
$m$, for any $f_{\boldsymbol{w}}$ of depth $d$ and a constant $C\left(B, f_{\boldsymbol{w}} \right)$ depending only on the maximal input norm $B$ and on the structure of $ f_{\boldsymbol{w}}$, we have

\begin{align}
L_{\gamma_{1}}\left(f_{\boldsymbol{w}}\right) & \le\hat{L}_{\gamma_{2}}\left(f_{\boldsymbol{w}}\right)+\text{\ensuremath{\mathcal{O}}}\left(\sqrt{\frac{C\left(B, f_{\boldsymbol{w}} \right)+\ln\frac{dm}{\delta}}{\left(\gamma_{2}-\gamma_{1}\right)^{2}m}}\right).
\end{align}
\end{theorem}
The proof is relegated to Sec. \ref{subsec_Neyshabur}. See Fig. \ref{fig_gamma_bound_test} for empirical corroboration of the bound above. \\

A common measure of the reconstruction performance of an AE is the squared-error loss
\begin{align} 
l_{SE}\left(x,\hat{x}\right) &=\norm{x-\hat{x}}_{2}^{2}={\displaystyle \sum_{j=1 }^{M}}\left(x\left[j\right]-\hat{x}\left[j\right]\right)^{2}.
\end{align}
We would like to be able to bound the generalization error in terms of this loss as well. Fortunately, we are able to bound $l_{SE}$ by a function of $l_{\gamma}$. 
Let 
\begin{align}
R\left(r,\gamma\right) & \triangleq rM+\left(1/2-\gamma\right)^{2}\left(1-r\right)M=rM\left(1-\left(1/2-\gamma\right)^{2}\right)+\left(1/2-\gamma\right)^{2}M.
\end{align}

\begin{lemma} Let $x$ be an input and $\hat{x}$ its reconstruction. Suppose that $l_{\gamma}\left(x,\hat{x}\right)$
is at most $r$. Then $l_{SE}\left(x,\hat{x}\right)$ is at most $R(r, \gamma)$.

\end{lemma}
Indeed, at most $rM$ entries are reconstructed with
accuracy less than $1/2-\gamma$. They contribute at most
$1\cdot rM$ to $l_{SE}$. The remaining $\left(1-r\right)\cdot M$
entries contribute at most $\left(1/2-\gamma\right)^{2}\left(1-r\right)M$
to $l_{SE}$, for a total loss at most
 $rM+\left(1/2-\gamma\right)^{2}\left(1-r\right)M$.
\begin{corollary} \label{cor_gamma_to_mu} By linearity of expectation, an expected $\gamma$-margin loss $L_{\gamma}\left(f\right)\le r$
implies a squared-error loss at most $R\left(r,\gamma\right)$. Similarly, by the Jensen inequality, the expected $L_{2}$ error
\begin{align}
\mu\left(f_{\boldsymbol{w}}\right) & \triangleq \mathbb{E}\left[\norm{f\left(x\right)-x}_{2}\right]
\end{align}
is bounded from above by $\sqrt{R\left(r,\gamma\right)}$. 
\end{corollary}
Suppose further that the reconstruction errors of the entries are distributed symmetrically around the average of the possible values. That is, that the distance from the corresponding input is, on average, $\nicefrac{\left((1/2-\gamma)^2+1^2\right)}{2}$ for the $rM$ entries with poor reconstruction, and $\nicefrac{(1/2-\gamma)^2}{2}$ for the $(1-r)M$ remaining entries. Then 

\begin{align} \label{eq_symmetric_r_mu_bound}
\mu\left(f_{\boldsymbol{w}}\right) & \le \sqrt{\frac{\left(\frac{1}{2}-\gamma\right)^{2}+1^{2}}{2}rM+\frac{\left(\frac{1}{2}-\gamma\right)^{2}}{2}\left(1-r\right)M}.
\end{align}
The empirical results in Fig. \ref{fig_r_mu_bound} suggest that this symmetric error assumption is reasonable. \\

 Substituting the generalization bound from Thm. \ref{thm_Neyshabur_AEs} into Corollary \ref{cor_gamma_to_mu}, we obtain the following bound for $\mu\left(f_{\boldsymbol{w}}\right)$.
  
\begin{theorem} \label{thm_mu_bound}
For any positive $0<\delta<1,0<\gamma_1 < \gamma_2<\frac{1}{2}$,
with probability at least $1-\delta$ over a training set of size
$m$, for any $f_{\boldsymbol{w}}$ of depth $d$ and network-related constant $C\left( f_{\boldsymbol{w}} \right)$ independent of $m$, we have
\begin{align}
\mu\left(f_{\boldsymbol{w}}\right) & \le\sqrt{R\left(\hat{L}_{\gamma_{2}}\left(f_{\boldsymbol{w}}\right)+\text{\ensuremath{\mathcal{O}}}\left(\sqrt{\frac{C\left( f_{\boldsymbol{w}} \right)+\ln\frac{dm}{\delta}}{\left(\gamma_{2}-\gamma_{1}\right)^{2}m}}\right), \gamma_1\right)}.
\end{align} \label{eq_mu_bound}
\end{theorem}
\subsection{Proof of Theorem. \ref{thm_Neyshabur_AEs}} \label{subsec_Neyshabur}
We follow the strategy appearing in \cite{Neyshabur_18}. \\

First, let us state the result in greater detail. Let $f_{\boldsymbol{w}}$ be an autoencoder with weights $\boldsymbol{w}=\left\{ W_{i}\right\} _{i=1}^{d}$ and ReLU non-linearities.  Let
\begin{align}
\norm{W}_2 & \triangleq \sup_{x\neq0} \frac{\norm{Wx}_2}{\norm{x}_2}, \nonumber \\
\norm{W}_F & \triangleq \sqrt{\sum_{i=1}^{p}\sum_{j=1}^{q}|w_{ij}|^2}
\end{align}
be the spectral and Frobenius norms of a $(p\times q)$-dimensional matrix $W$.
 Let $B$ be the maximum $L_{2}$
norm of an input, $d$ the depth of $f$, $h$ the upper bound
on the number of output units in each layer. 
\begin{theorem} (Detailed version of Thm. \ref{thm_Neyshabur_AEs}) \label{thm_detailed_AE_bound}
For any $B,d,h>0$ and any $0<\delta<1,0<\gamma_1 < \gamma_2<1/2$,
with probability at least $1-\delta$ over a training set of size
$m$, for any $\boldsymbol{w}$, we have
\begin{align}
L_{\gamma_1}\left(f_{\boldsymbol{w}}\right) & \le\hat{L}_{\gamma_2}\left(f_{\boldsymbol{w}}\right)+\text{\ensuremath{\mathcal{O}}}\sqrt{\frac{B^{2}d^{2}h\ln\left(dh\right)\Pi_{i=1}^{d}\text{\ensuremath{\norm{W_{i}}}}_{2}^{2}\sum_{i=1}^{d}\frac{\text{\ensuremath{\norm{W_{i}}}}_{F}^{2}}{\text{\ensuremath{\norm{W_{i}}}}_{2}^{2}}+\ln\frac{dm}{\delta}}{(\gamma_2-\gamma_1)^{2}m}}.
\end{align}
\end{theorem}
The proof consists of three steps. Firstly, we show that a small perturbation of the weight matrices implies a small perturbation of the network output (Lemma \ref{lem_perturbation_bound}, Lemma 2 in \cite{Neyshabur_18}). Secondly, for perturbations $\boldsymbol{u}$ of the network parameters such that the network output does not change much relative to $\gamma_2-\gamma_1$, we state a PAC-Bayesian bound controlling $L_{\gamma_1}(f)-\hat{L}_{\gamma_2}(f)$ by means of $\boldsymbol{w}+\boldsymbol{u}$ (Lemma \ref{lem_PAC_Bayes}, analogous to Lemma 1 in \cite{Neyshabur_18}). Finally, we use Lemma \ref{lem_perturbation_bound} to calculate the maximal amount of perturbation that satisfies the conditions of Lemma \ref{lem_PAC_Bayes}. This level of perturbation, substituted into the PAC-Bayesian bound, yields the theorem. 

\begin{lemma} (Perturbation bound) \label{lem_perturbation_bound}
Let $\boldsymbol{u}=\left\{ U_{i}\right\} _{i=1}^{d}$ be a perturbation such that $\left\Vert U_{i}\right\Vert _{2}\le\frac{1}{d}\left\Vert W_{i}\right\Vert _{2}$. Then for any input $x$,
\begin{align}
\left|f_{\boldsymbol{w}+\boldsymbol{u}}\left(x\right)-f_{\boldsymbol{w}}\left(x\right)\right|_{2} & \le eB\left(\prod_{i=1}^{d}\left\Vert W_{i}\right\Vert _{2}\right)\sum_{i=1}^{d}\frac{\left\Vert U_{i}\right\Vert _{2}}{\left\Vert W_{i}\right\Vert _{2}}.
\end{align}
\end{lemma}
Recall that the \emph{Kullback-Leibler divergence} between two distributions $P$
and $Q$  is
\begin{align}
KL\left(Q\parallel P\right) & \triangleq \mathbb{E}_{Q}\left[\ln\frac{Q}{P}\right].
\end{align}

\begin{lemma} (PAC-Bayesian bound) \label{lem_PAC_Bayes}
Let $f_{\boldsymbol{w}}:\mathcal{X}\rightarrow\mathcal{X}$ be
an autoencoder, $P$ a data-independent distribution on the parameters.
Then for any $0<\gamma_1<\gamma_2<1/2,0<\delta<1$, w.p. at least $1-\delta$, for any random perturbation $\boldsymbol{u}$ s.t.
$\mathbb{P}_{\boldsymbol{u}}\left[{\displaystyle \max_{x\in\mathcal{X}}}\left|f_{\boldsymbol{w}+\boldsymbol{u}}\left(x\right)-f_{\boldsymbol{w}}\left(x\right)\right|_{\infty}<\frac{(\gamma_2-\gamma_1)}{4}\right]\ge1/2$,
we have
\begin{align} \label{eq_PAC_Bayes_bound}
L_{\gamma_1}\left(f_{\boldsymbol{w}}\right) & \le\hat{L}_{\gamma_2}\left(f_{\boldsymbol{w}}\right)+4\sqrt{\frac{KL\left(\boldsymbol{w}+\boldsymbol{u}\parallel P\right)+\ln\frac{6m}{\delta}}{m-1}}.
\end{align}
\end{lemma}
Let $\beta=\left(\prod_{i=1}^{d}\left\Vert W_{i}\right\Vert _{2}\right)^{\frac{1}{d}}$.
Consider the weights $\tilde{W}_{i}=\frac{\beta}{\left\Vert W_{i}\right\Vert _{2}}W_{i}$.
By the homogeneity of ReLU, $f_{\tilde{\boldsymbol{w}}}=f_{\boldsymbol{w}}$.
Also, $\prod_{i=1}^{d}\left\Vert W_{i}\right\Vert _{2}=\prod_{i=1}^{d}\left\Vert \tilde{W}_{i}\right\Vert _{2}$
and $\frac{\left\Vert W_{i}\right\Vert _{F}}{\left\Vert W_{i}\right\Vert _{2}}=\frac{\left\Vert \tilde{W}_{i}\right\Vert _{F}}{\left\Vert \tilde{W}_{i}\right\Vert _{2}}$. We can, therefore, assume that all weights are normalized and $\left\Vert W_{i}\right\Vert _{2}=\beta$
for all $i$. \\

Consider $\boldsymbol{u} \sim \mathcal{N}(0, \sigma^2)$ and a prior distribution $P$ of the same form. By Thm. 4.1 in \cite{Tropp_12}, with probability at least 1/2,
\begin{align} \label{eq_bound_Ui}
\norm{U_i}_2 & \le \sigma\sqrt{2h\ln (4dh)}.
\end{align} 
By Lemma \ref{lem_perturbation_bound}, for an appropriate $\sigma$,
\begin{align}
\max\left|f_{\boldsymbol{w}+\boldsymbol{u}}\left(x\right)-f_{\boldsymbol{w}}\left(x\right)\right| & \le eB\beta^{d}\sum_{i}\frac{\left\Vert U_{i}\right\Vert _{2}}{\beta} \nonumber \\
 & \le eB\beta^{d-1}\sigma\sqrt{2h\ln\left(4dh\right)} \nonumber \\
& \le\frac{(\gamma_2-\gamma_1)}{4}.
\end{align}
For such a $\sigma$, $\boldsymbol{u}$ satisfies the condition of Lemma \ref{lem_PAC_Bayes}. We can now bound the $KL$ term in the PAC-Bayesian bound for the chosen $P$ and $\boldsymbol{u}$\footnote{We have skipped over a nuance necessary to ensure that the prior $P$ is data-independent. See the end of the proof of Theorem 1 in \cite{Neyshabur_18} for the details},
\begin{align} \label{eq_KL_bound}
KL\left(\text{\ensuremath{\boldsymbol{w}}+\ensuremath{\boldsymbol{u}}}\parallel P\right) & \le \frac{\norm{\boldsymbol{w}}^2}{2\sigma^2} \le \text{\ensuremath{\mathcal{O}}}\sqrt{\frac{B^{2}d^{2}h\ln\left(dh\right)\Pi_{i=1}^{d}\text{\ensuremath{\norm{W_{i}}}}_{2}^{2}\sum_{i=1}^{d}\frac{\text{\ensuremath{\norm{W_{i}}}}_{F}^{2}}{\text{\ensuremath{\norm{W_{i}}}}_{2}^{2}}}{(\gamma_2-\gamma_1)^{2}}}.
\end{align}
Substituting Eq. \ref{eq_KL_bound} into Eq. \ref{eq_PAC_Bayes_bound} completes the proof.
\begin{remark}
Note that the upper bound on $\norm{U_i}_2$ given in Eq. \ref{eq_bound_Ui} depends on the dimensions of $U_i$. Thm. \ref{thm_detailed_AE_bound} and its proof simply use $h$, the largest output unit number of any layer. Assuming that the layer sizes decrease exponentially approaching the bottleneck (see, e.g., \cite{Hinton_Salakhutdinov_2006}), there is some room for tightening the bound.
\end{remark}

\section{Autoencoders and Semi-Supervised Learning} \label{sec_AEs_SSL}

\begin{figure}[ht] 
\begin{center}
\includegraphics[totalheight=5.0cm]{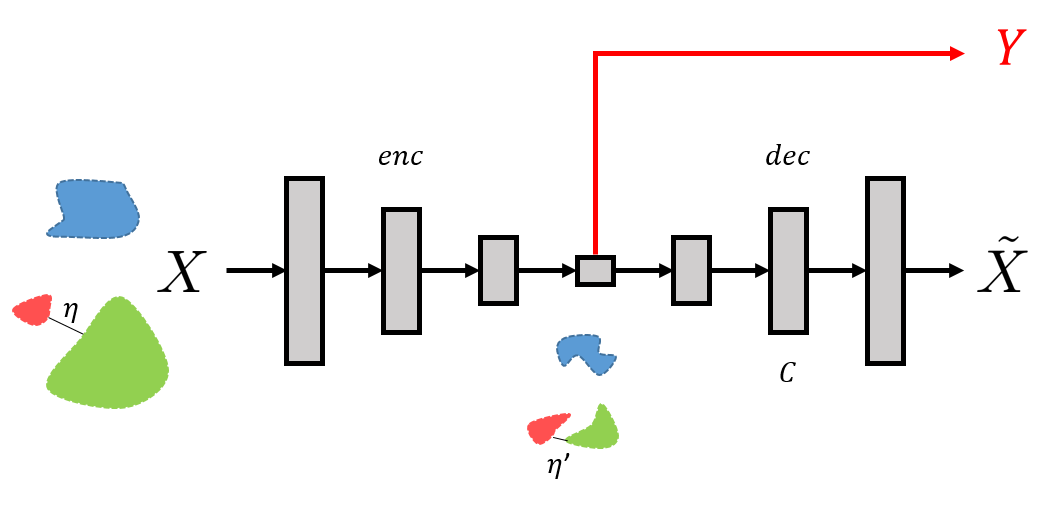}

\caption{An autoencoder with a semi-supervised learner applied to the encoder output. The Lipschitz constant of $dec$ is denoted by $C$. The input distribution satisfies the clustering assumption with cluster-margin $\eta$. The encoder distorts the input distribution, but, for a sufficiently good autoencoder, the distribution at the AE bottleneck still satisfies the clustering assumption with cluster-margin $\eta ' > 0$. }\label{fig_AE_with_supervised}
\end{center}
\end{figure}

In this section we show that, under appropriate assumptions, a sufficiently good autoencoder can contribute to the advantage of SSL over any supervised learning scheme. Specifically, we consider the following strategy - first training the AE on the unlabeled data, and then applying the bound in Prop. \ref{prop_Singh} to the code, that is, to the output of the encoder (see Fig. \ref{fig_AE_with_supervised}). We stress that we do not propose this strategy as an optimal empirical approach. Indeed, training to minimize both reconstruction and supervised losses simultaneously has been established as a more successful approach, in practice (e.g., \cite{DSN}, \cite{JAE_18}.) However, the scheme we are considering allows for a theoretical treatment and for an explanation of the relationship between the autoencoder performance and its contribution to semi-supervised learning.

We need some further notation, in order to state our main result in this section. Denote by $G_{\epsilon}\left(f_{\boldsymbol{w}}\right)\subset\mathcal{X}$\footnote{We will occasionally omit $f$ or $f_{\boldsymbol{w}}$ and simply write $G_{\epsilon}$.}
the subset $\left\{ x:\norm{f_{\boldsymbol{w}}\left(x\right)-x}_{2}-\mu\left(f_{\boldsymbol{w}}\right)<\epsilon\right\} $,
that is, the inputs for which the reconstruction error deviates from
$\mu\left(f_{\boldsymbol{w}}\right)$ by at most $\epsilon$. Note that by the Markov inequality, for $x \in\mathcal{X}$,
\begin{align} \label{eq_measure_G_eps}
P\left(\norm{f_{\boldsymbol{w}}\left(x\right)-x}_{2}-\mu\left(f_{\boldsymbol{w}}\right)>\epsilon\right)=P\left(x\notin G_{\epsilon}\left(f_{\boldsymbol{w}}\right)\right) & \le\frac{\mu\left(f_{\boldsymbol{w}}\right)}{\epsilon},
\end{align}
or in other words, the measure of $G_{\epsilon}\left(f_{\boldsymbol{w}}\right)$ is
at least $1-\mu \left(f_{\boldsymbol{w}}\right)/ \epsilon$. Note that this allows us to trade off the
measure of $G_{\epsilon}$ for the tightness of $\epsilon$ (see Fig. \ref{fig_G_eps}). Observe,
too, that by Thm. \ref{eq_mu_bound},  $\mu\rightarrow0$ as $m\rightarrow\infty,\hat{L}_{\gamma_{2}}\rightarrow0$
and $\gamma_{1}\rightarrow\gamma_{2}$. Thus, as the generalization
error of $f$ vanishes, so does the set of ``bad'' inputs. Denote by $\mathcal{D}_{G_{\epsilon}}$ the distribution induced by $\mathcal{D}$ on $G_\epsilon$, and by $\mathcal{D}_{enc(G_{\epsilon})}$ the corresponding distribution on $enc(G_{\epsilon})$.

\begin{theorem} \label{thm_Singh_AEs} Assume that the input distribution $\mathcal{D}$ satisfies the clustering assumption with margin $\eta$. Let $f$ be an autoencoder with expected $L_2$ reconstruction loss $\mu(f)$, bottleneck dimension $N_b<N$ and decoder Lipschitz constant $C$\footnote{The decoder is a Lipschitz function. Indeed, $C$ is at most the product of the spectral norms of the weight matrices in $dec$, though that is typically a very loose bound. See \cite{Arora_Compression_18} for a discussion of the behavior of $C$.}. Then for any $\epsilon>0$, $\mathcal{D}_{enc(G_{\epsilon})}$ satisfies the clustering assumption with cluster-margin at least
\begin{align}\label{eq_eta'}
\eta'=\nicefrac{\left(\eta-2\left(\mu(f)+\epsilon\right)\right)}{C}.
\end{align} Furthermore, suppose there exists a clairvoyant learner  $\mathcal{A}_{\mathcal{D}_{enc(G_{\epsilon})},n}$ for which
\begin{align}
 \mathbb{E}[\mathcal{E}(\mathcal{A}_{\mathcal{D}_{enc(G_{\epsilon})},n})] \le \varepsilon(n).
\end{align}
Then there exists a semi-supervised learner $\mathcal{A}_{m,n}$ such that if $\eta' > C_0 ((\log m)^2 / m)^{1/N_b}$, then
\begin{align}
\mathbb{E}[\mathcal{E}(\mathcal{A}_{m,n})] \le \varepsilon(n) + O\left( \frac{1}{m}+n \left( \frac{(\log m)^2}{m} \right) ^{1/N_b} \right). \label{eq_Singh_bound_AEs}
\end{align}
The constant $C_0$ does not depend on $\eta$, $m$ or $n$.
\end{theorem} 
Before proving the theorem, we observe that whenever $\nicefrac{\eta}{\eta'}<\log(N-N_b)$, both the condition on $m$ and the bound in Eq. \ref{eq_Singh_bound_AEs} are improved relative to Prop. \ref{prop_Singh}.

~
	
Now, consider two
input points $x,y\in G_{\epsilon}\left(f\right)$. Applying
a standard ``$3$ epsilon'' argument,
\begin{align}
\norm{x-y}_{2} & \le\norm{f\left(x\right)-x}_2+\norm{f\left(x\right)-f\left(y\right)}_{2}+\norm{f\left(y\right)-y}_{2},\nonumber \\
 & \le \norm{f\left(x\right)-f\left(y\right)}_{2}+2\left(\mu\left(f\right)+\epsilon\right).
\end{align}
In particular, for $x,y$ at least $\eta$ apart, $\norm{f\left(x\right)-f\left(y\right)}_{2}$
is at least $\eta-2\left(\mu(f)+\epsilon\right)$.

~

We have established that if an autoencoder generalizes well,
it does not bring two input points in $G_{\epsilon}\left(f\right)$
too close together. Recall that $C$ is the Lipschitz constant of the decoder.
If, for any points $x,y$, $\norm{f\left(x\right)-f\left(y\right)}_{2}$
is at least some $d$, then $\norm{enc\left(x\right)-enc\left(y\right)}_{2}$
cannot be less than $\nicefrac{d}{C}$. This implies that $enc$ maps
clusters at least $\eta$ apart to clusters at least $\eta'=\nicefrac{\eta-2\left(\mu\left(f\right)+\epsilon\right)}{C}$
apart. In other words, if $\mathcal{D}$, the input distribution, satisfies the clustering
assumption with margin $\eta$, then its restriction $\mathcal{D}_{G_{\epsilon}}$ does as well, and $\mathcal{D}_{enc(G_{\epsilon})}$, the output distribution of $enc$, satisfies the clustering assumption with cluster-margin $\eta'$. Applying Prop. \ref{prop_Singh} to the $\mathcal{D}_{enc(G_{\epsilon})}$ completes the proof.

\section{Experiments} \label{sec_experiments}
\begin{figure}[ht!]
\centering     
\subfigure[]{\label{fig_gamma_bound_test}\includegraphics[width=50mm]{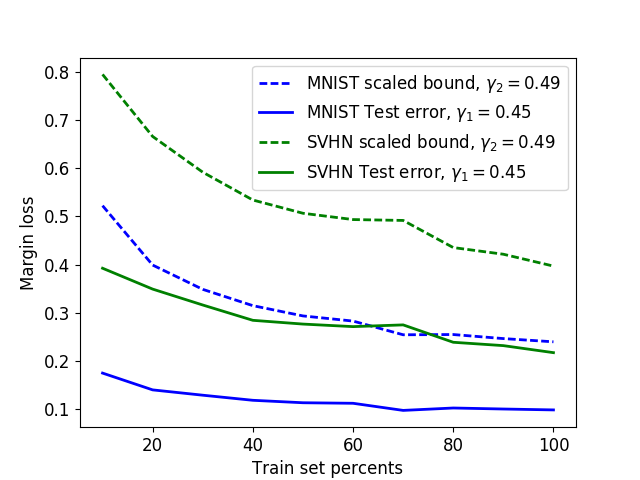}}
\subfigure[]{\label{fig_r_mu_bound}\includegraphics[width=50mm]{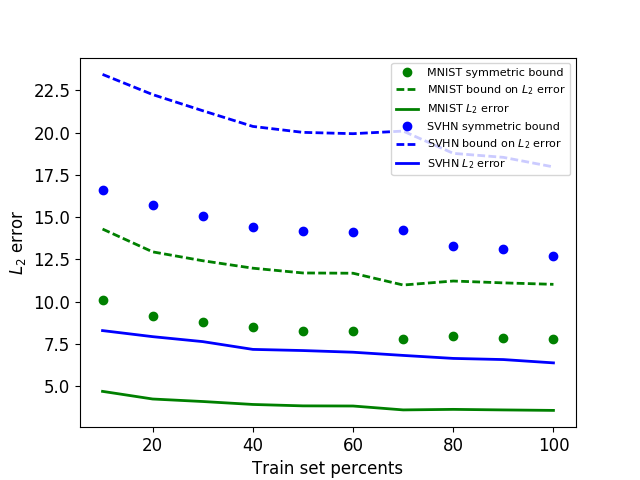}}
\subfigure[]{\label{fig_G_eps}\includegraphics[width=50mm]{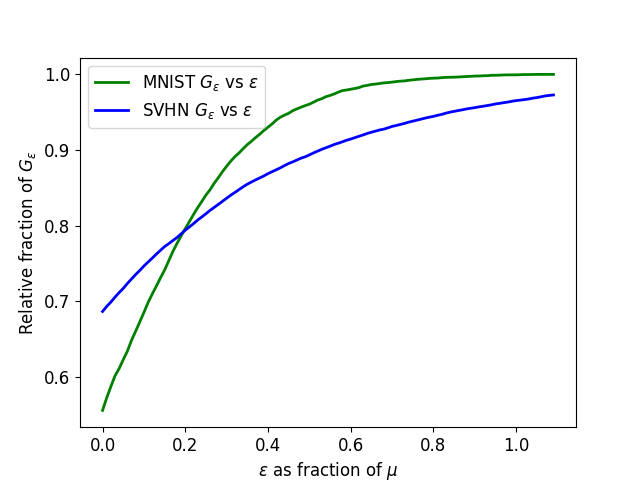}}
\caption{(a) The (scaled-down) generalization bound in Thm. \ref{thm_Neyshabur_AEs} correlates well with the empirical test error as the sample size increases from $10\%$ of the training set to $100\%$. (b) The bound for $\mu(f)$ as a function of $L_\gamma$ given in Corollary \ref{cor_gamma_to_mu} correlates well with the empirical $L_2$ error as the the sample size increases from $10\%$ of the training set to $100\%$. The worst-case bound is overly pessimistic, but the bound in Eq.  \ref{eq_symmetric_r_mu_bound} derived under the symmetric-error assumption is much closer to reality. (c) $G_{\epsilon}$, the subset of samples with $L_2$ reconstruction error at most $\mu(f)+\epsilon$, rapidly becomes almost all of the test set as $\nicefrac{\epsilon}{\mu(f)}$ increases to $1$.}
\end{figure}

All experiments were implemented in Keras (\cite{chollet2015}) over
Tensorflow (\cite{tensorflow2015-whitepaper}). We use two digit image datasets for our experiments. The MNIST dataset (\cite{MNIST}) is a collection of 70000 grayscale $28\times28$ images of hand-written digits, split into 60,000 training and 10,000 test samples. The SVHN dataset (\cite{SVHN})  is a collection of 99289 $32\times32\times3$ RGB images of hand-written digits, split into 73,257 training and 26,032 test samples. We have the converted the SVHN samples into grayscale. First, we provide evidence for the generalization bound in Thm. \ref{thm_Neyshabur_AEs}. For each dataset, we train an autoencoder on an increasing fraction of the training set, and plot the bound (divided by the constant $C(B, f_{\boldsymbol{W}})$ vs. the empirically observed test error (Fig \ref{fig_gamma_bound_test}). The margin values we use are $\gamma_1=0.45, \gamma_2=0.49$. We see that, for both datasets, the bound correlates well with the test error. Moreover, the plot trends suggest an asymptotic convergence of the bound to the test error. 

Next, we examine the control over $\mu(f)$ as a function of $l_{\gamma}$ that Corollary \ref{cor_gamma_to_mu} provides. Fig. \ref{fig_r_mu_bound} plots the empirical average $L_2$ reconstruction error over the test set vs the predicted bound. The worst-case bound $\sqrt{R(l_{\gamma}, \gamma)}$ correlates well with the empirical $L_2$ error, but it is overly pessimistic by a factor of approximately 3. The average-case bound in Eq.  \ref{eq_symmetric_r_mu_bound} is closer to the observed error, loose only by a factor of approximately $2$.

The proof of Thm. \ref{thm_Singh_AEs} requires a restriction to $G_{\epsilon}$, the set of samples with small reconstruction error. A reasonable concern is that such a restriction rejects a large fraction of the inputs. While Eq. \ref{eq_measure_G_eps} provides some guarantees on the size of $G_{\epsilon}$ for negligible $\mu(f)$-s, Fig. \ref{fig_G_eps} shows that, already for $\epsilon$ values small relative to $\mu(f)$, most test samples are in $G_{\epsilon}$.

Finally, in Table. \ref{table_ae_SSL_results} we examine the various quantities appearing in Thm. \ref{thm_Singh_AEs}. The first and second rows use a small and a large autoencoder, respectively, trained on MNIST. The third row uses an autoencoder trained on SVHN. The first column, $N\rightarrow N_b$, describes the change in dimensions from the AE input to the bottleneck. The second column, $\eta$, gives the estimated cluster-margin of the input distribution. $\eta'$ is the estimated cluster-margin of the bottleneck distribution. $C$ is the estimated decoder Lipschitz constant. Finally, the fifth column gives the extent to which the $\nicefrac{\log(m)^2}{m}^{\nicefrac{1}{N_b}}$ term in Eq. \ref{eq_Singh_bound_AEs} improves due to the dimension reduction, for the corresponding value of $m$. We can see that the change in the cluster-margin is roughly inverse to $C$ (though, for the given training sets, $\mu(f)$ was not small enough for Eq. \ref{eq_eta'} to yield positive values of $\eta'$).

\begin{center} \begin{table}[ht] \caption{Autoencoder Lipschitz Constant, Cluster-Margin and SSL Bound }  
\begin{centering}  \label{table_ae_SSL_results}   \begin{tabular}{llllll} \toprule 
				 & $N\rightarrow N_b$      & $\eta$ & $\eta '$& $C$        & Bound improvement    \\ \midrule  
MNIST                        & $784 \rightarrow 30$   & $3.99$ & $1.60$ & $3.39$    & $1.22$ \\
MNIST (large network) & $784 \rightarrow 50$   & $3.99$ & $3.41$ & $1.86$    & $1.12$ \\
SVHN                          & $1024 \rightarrow 50$ & $1.58$ & $0.12$ &  $24.93$ & $1.23$ \\
\tabularnewline \bottomrule    \end{tabular} \par\end{centering}  \vspace*{4mm} \caption*{$N\rightarrow N_b$ denotes the dimension reduction the AE encoder performs. $\eta$ is the empirical input cluster-margin. $\eta'$ is the empirical cluster-margin at the AE code. $C$ is the estimated Lipschitz constant of the decoder. Bound improvement refers to the (multiplicative) improvement in the $\nicefrac{\log(m)^2}{m}^{\nicefrac{1}{N_b}}$ term in Eq. \ref{eq_Singh_bound_AEs}. The decrease in $\eta'$ is roughly inverse to $C$, as predicted by Eq. \ref{eq_eta'}.}  \end{table} \par\end{center}
\section{Conclusion}
We have adapted existing generalization bounds for feedforward networks, together with a novel reconstruction loss, to obtain a generalization bound for autoencoders. To the best of our knowledge, this is the first such bound. We went on to tie the good reconstruction performance of an autoencoder to a non-contractiveness property of the encoder component. This property, in turn, implies the ability to trade off cluster-margins between input clusters for reduced dimension, which is beneficial for semi-supervised learning. Empirical evidence supports our theoretical results.

The bound we have obtained concerns only the \emph{gap} between the empirical and expected losses. It neither guarantees the existence of an autoencoder achieving a negligible empirical error nor explains why such networks seem to exist in practice, particularly for images. We believe that the answer has to do with the properties of natural images - that typical image datasets satisfy a \emph{manifold hypothesis}. That is, they lie on, or near, a low-dimensional manifold that is mapped to a higher dimension where they are observed. Assuming the mapping is invertible, and both the mapping and its inverse can be approximated well by sufficiently expressive networks, this does imply the existence of a good autoencoder for the dataset. Such considerations lead us to believe that a good generative model for the data (possibly along the lines of \cite{Patel_18}) could shed further light on unsupervised and semi-supervised learning with autoencoders. 

An interesting and worthwhile extension of our work would be to consider more practical approaches to SSL. Specifically, combining supervised and unsupervised losses through shared layers, as is often done in practice. Such approaches have been shown to be effective both in SSL and transfer learning, and the present approach could shed theoretical light on their success. 

\acks{We thank Ron Amit for numerous useful suggestions and corrections.}

\pagebreak

\bibliography{COLT19_bibtex_db}






\end{document}